\documentclass{article} 
\usepackage{myarxiv,graphicx,color, natbib}
\usepackage[usenames,dvipsnames]{xcolor}

\date{}
\usepackage{url,hyperref,lineno,microtype,subcaption,amsmath}
\usepackage{tabularx}

\author{Martino Sorbaro,$^{1,2}$ Qian Liu,$^{1}$ Massimo Bortone,$^{1}$ and Sadique Sheik$^{1, }$\thanks{\url{sadique.sheik@aictx.ai}} \\
$^{1}$aiCTX AG, Z\"urich, Switzerland, \\$^{2}$ Institute of Neuroinformatics, Univ. of Z\"urich and ETH Z\"urich, Switzerland}

\begin{document}
\onecolumn

\title{Optimizing the energy consumption of spiking neural networks for neuromorphic applications}



\maketitle

\begin{abstract}

In the last few years, spiking neural networks have been demonstrated to perform on par with regular convolutional neural networks. 
Several works have proposed methods to convert a pre-trained CNN to a Spiking CNN without a significant sacrifice of performance.
We demonstrate first that quantization-aware training of CNNs leads to better accuracy in SNNs. 
One of the benefits of converting CNNs to spiking CNNs is to leverage the sparse computation of SNNs and consequently perform equivalent computation at a lower energy consumption. Here we propose an optimization strategy to train efficient  spiking networks with lower energy consumption, while maintaining similar accuracy levels. We demonstrate results on the MNIST-DVS and CIFAR-10 datasets.



\end{abstract}


\section{Introduction}

Since the early 2010s, computer vision has been dominated by the introduction of convolutional neural networks (CNNs), which have yielded unprecedented success in previously challenging tasks such as image recognition, image segmentation or object detection, among others. Considering the theory of neural networks was mostly developed decades earlier, one of the main driving factors behind this evolution was the widespread availability of high-performance computing devices and general purpose Graphic Processing Units (GPU). In parallel with the increase in computational requirements \citep{strubell2019energy}, the last decades have seen a considerable development of portable, miniaturized, battery-powered devices, which pose constraints on the maximum power consumption.

Attempts at reducing the power consumption of traditional deep learning models have been made. Typically, these involve optimizing the network architecture, in order to find more compact networks (with fewer layers, or fewer neurons per layer) that perform equally well as larger networks. One approach is energy-aware pruning, where connections are removed according to a criterion based on energy consumption, and accuracy is restored by fine-tuning of the remaining weights \citep{yang2017designing, molchanov2016pruning}. Other work looks for more efficient network structures through a full-fledged architecture search \citep{cai2018proxylessnas}. The latter work was one of the winners of the Google ``Visual Wake Words Challenge'' at CVPR 2019, which sought models with memory usage under 250 kB, model size under 250 kB and per-inference multiply-add count (MAC) under 60 millions.

Using spiking neural networks (SNNs) on neuromorphic hardware is an entirely different, and much more radical, approach to the energy consumption problem. In SNNs, like in biological neural networks, neurons communicate with each other through isolated, discrete electrical signals (spikes), as opposed to continuous signals, and work in continuous instead of discrete time. Neuromorphic hardware \citep{indiveri2011neuromorphic,furber2016large,thakur2018large,esser2016convolutional} is specifically designed to run such networks with very low power overhead, with electronic circuits that faithfully reproduce the dynamics of the model in real time, rather than simulating it on traditional von Neumann computers. Some of these architectures (including Intel's Loihi, IBM's TrueNorth, and aiCTX's DynapCNN) support convolution operations, which are necessary for modern computer vision techniques, by an appropriate weight sharing mechanism.

The challenge of using SNNs for machine learning tasks, however, is in their training. Mimicking the learning process used in the brain's spiking networks is not yet feasible, because neither the learning rules, nor the precise fitness functions being optimized are sufficiently well understood, although this is currently a very active area of research \citep{marblestone2016toward, richards2019deep}. Supervised learning routines for spiking networks have been developed \citep{bohte2002error,mostafa2017supervised,nicola2017supervised,neftci2019surrogate,shrestha2018slayer}, but are slow and challenging to use. For applications which have little or no dependence on temporal aspects, it is more efficient to train an analog network (i.e., a traditional, non-spiking one) with the same structure, and transfer the learned parameters onto the SNN, which can then operate through rate coding. In particular, the conversion of pre-trained CNNs to SNNs has been shown to be a scalable and reliable process, without much loss in performance \citep{diehl2015fast,rueckauer2017conversion,sengupta2019going}. But this approach is still challenging, because the naive use of analog CNN weights does not take into account the specific characteristics and requirements of SNNs. In particular, SNNs are more sensitive than analog networks to the magnitude of the input. Naive weight transfer can, therefore, lead to a silent SNN, or, conversely, to one with unnecessarily high firing rates, which have a high energy cost.

Here, we propose a hybrid training strategy which maintains the efficiency of training analog CNNs, while accounting for the fact that the network is being trained for eventual use in SNNs. Furthermore, we include the energy cost of the network's computations directly in the loss function during training, in order to minimize it automatically and dynamically. We demonstrate that networks trained with this strategy perform better per Joule of energy utilized. While we demonstrate the benefit of optimizing based on energy consumption, we believe this strategy is extendable to any approach that uses back-propagation to train the network, be it through a spiking network or a non-spiking network.


In the following sections, we will detail the training techniques we devised and applied for these purposes. We will test our networks on two standard problems. The first is the MNIST-DVS dataset of Dynamic Vision Sensor recordings. DVSs are event-based sensors, and, as such, the analysis of their recordings is an ideal application of spike-based neural networks. The second is the standard CIFAR-10 object recognition benchmark, which provides a reasonable comparison on computation cost to non-spiking networks. For each of these tasks, we will demonstrate the energy-accuracy trade-off of the networks trained with our methods. We show that significant amounts of energy can be saved with a small loss in performance, and conclude that ours is a viable strategy for training neuromorphic systems with a limited power budget.

\section{Materials and Methods}

In most state-of-the art neuromorphic architectures with time multiplexed units like~\cite{Merolla_2014, davies2018loihi, furber2014spinnaker}, the various states need to be fetched from memory and rewritten. Such operations happen every time a neuron receives a synaptic event. Whenever one of these operations is performed, the neuromorphic hardware consumes a certain amount of energy. For instance in \cite{indiveri2019importance} the authors show that this energy consumption is usually of the order of $10^{-11}$ J. 
While there are several other processes that consume power on a neuromorphic device, the bulk of the active power on these devices is used by the synaptic operations.
Reducing their number is therefore the most natural way to keep energy usage low.

In this paper we explore strategies to lower synaptic operations and evaluate their effect on the network's computational performance. We suggest to train, or fine-tune, networks with an additional loss term which explicitly enforces lower activations in the trained network --- and consequently lower firing rates of the corresponding spiking network. This is analogous to the $L_1$ term used by \cite{esser2016convolutional} and \cite{neil2016learning}, but applied on synaptic operations directly rather than firing rates, and set up so that a target SynOp count value can be set. Additionally, we introduce quantization of the activations on each layer, which mimics the discretization effect of spiking networks, so that the network activity remains at reasonable levels even when the regularization term is strong. The following sections illustrate the technical details and introduce the datasets and networks we use for evaluation.

\subsection{Training strategies}

\subsubsection{Parameter scaling}

By scaling the weights, biases and/or thresholds of neurons in different layers, we can influence the number of spikes generated in each layer, thereby allowing us to tune the synaptic activity of the model. This is easy to do, even with pre-trained weights. For a scale-invariant network, such as any network whose only non-linearities are ReLUs, this method attains perfect results, because a linear rescaling of the weights causes a linear rescaling of the output, which gives identical results for classification tasks where we select the class that receives the highest activation.

We use this method as a baseline comparison for our results. We chose to rescale the weights of the first convolutional layer of our network by a variable factor $\rho$:
\[ w_0' = \rho w_0, \]
which is equivalent to a rescaling of the input signal by the same factor. Note that an increase/decrease in the first layer's output firing rate causes a correspondent increase/decrease in the activation of all the subsequent layers, and thereby reduces the global energy consumption of the whole network.

For baseline comparisons, we also apply the ``robust'' weight scaling suggested by \cite{rueckauer2017conversion}. This consists of a per-layer scaling of weights, in such a way that the maximum level of activation is constant along the network. For robustness, the 99th percentile of activations is taken as a measure of output magnitude in each layer, estimated from forward passes over 25600 samples of the training set. In this way, the activity of the network is balanced over its layers, in the sense that no layer unnecessarily amplifies or reduces the activity level compared to its input.

The scale-invariance property of ReLU functions does not hold for the corresponding spiking network, and small activation values could cause discretization errors, or even yield a completely silent spiking network from a perfectly functional analog network.

\subsubsection{Synaptic Operation optimization}

We measure the activity of the network, for each layer group, in correspondence with the ReLU operations, which effectively correspond to the spikes from an equivalent SNN (Supplementary Figure 2). We denote the activity of neuron $i$ in layer $\ell$ as $a_i^\ell$. We define the \emph{fan-out} of each group of layers, $f_\text{out}^\ell$, as the number of units of layer $\ell + 1$ that receive the signal emitted by a single neuron in layer $\ell$. This measure is essential in estimating the number of synaptic operations~(SynOps) $s^\ell$ elicited by each layer:
\begin{equation}
    s^\ell = f_\text{out}^\ell \sum_i a_i^\ell
\end{equation}

We directly add this number to the loss we want to minimize, optionally specifying a target value $S_0$ for the desired number of SynOps:
\begin{equation}\label{eq:synoploss}
    \mathcal{L} = \mathcal{C}(a^\text{output}, t) + \alpha \left({S_0 - \sum_\ell s^\ell} \right)^2
\end{equation}
where $\mathcal{C}$ is the cross-entropy loss, $t$ is the target label, and $\alpha$ is a constant. We will refer to this additional term as \emph{SynOp loss}. In this work, we will always choose $\alpha = 1/S_0^2$, in order to keep the SynOp loss term normalized independently of $S_0$. Although setting $S_0 = 0$ and tweaking the value of $\alpha$ instead is also a valid choice, we found it easier to set a direct target for the power budget, which leads to more predictable results.

Additionally, we performed some experiments where an $L_1$ penalty on activations was used, without fanout-based weighting. Against our expectations, we did not find a significant difference in power consumption between the models trained with or without per-layer weighting (Supplementary Figure 1). However, we use the fanout-based penalty throughout this paper, since this addresses the power consumption more directly, and we cannot rule out that this difference may be more significant in larger networks.

\subsubsection{Quantization-aware training and surrogate gradient}

Optimizing for energy consumption with the SynOp loss mentioned above has unintended consequences. During training, the optimizer tries to achieve smaller activations, but cannot account for the fact that, when the activations are too small, discretization errors become more prominent. Throughout this paper, by discretization error we mean the discrepancy that occurs when a real number needs to be represented in a discrete way --- namely, the value of each neuron's activation, which is continuous, needs to be translated in a finite number of spikes, leading to inevitable approximations. To solve this issue, we introduce a form of quantization during training.
The quantization of activations mimics, in the context of an analog network, a form of discretization analogous to what happens in a spiking network. Therefore, the network can be already aware of the discretization error at training time, and automatically adjust its parameters in order to properly account for it.
To this end, we turn all ReLU activation functions into ``quantized'' (i.e. step-wise) ReLUs, which additionally truncate the inputs to integers, as follows:
\begin{equation}
    \operatorname{QReLU}(x) = \begin{cases} 0 & x \leq 0  \\ \lfloor x\rfloor & x>0 \end{cases}
\end{equation}
where $\lfloor \cdot \rfloor$ indicates the floor operation.
This choice introduces a further problem: this function is discontinuous, and its derivative is uniformly zero wherever it is defined. To avoid the zeroing of gradients during the backward pass, we use a \emph{surrogate gradient} method \citep{neftci2019surrogate}, whereby the gradient of QReLU is approximated with the gradient of a normal ReLU during the backward pass:
\begin{equation}
    \nabla_x \operatorname{QReLU}(x) \approx \begin{cases} 0 & x \leq 0  \\ 1 & x>0 \end{cases}    
\end{equation}
This is not the only way to approximate the gradient of a step-wise function in a meaningful way, and closer approximations are certainly possible; however, we found that this linear approximation works sufficiently well for our purposes.

In this work, we apply QReLUs in combination with the SynOp loss term illustrated in the previous section, but quantization on activations could be independently used for a more accurate training of spiking networks. We note that quantization-aware training in different forms has been used before \citep{guo2018survey, hubara2017quantized}, but its typical purpose is to sharply decrease the memory consumption of CNNs, by storing both activations and weights as lower-precision numbers (e.g. as int8 instead of the typical float32). PyTorch recently started providing support utilities for this purpose.\footnote{\url{https://pytorch.org/docs/stable/quantization.html}}

\subsection{Spiking network simulations with Sinabs}
After training, we tested our trained weights on spiking network simulations. Unlike tests done on analog networks, these are time-dependent simulations, which fully account for the time dynamics of the input spike trains, and closely mimic the behaviour of a neuromorphic hardware implementation, like DynapCNN~\citep{liu2019live}. Our simulations are written using the Sinabs Python library\footnote{\url{https://aictx.gitlab.io/sinabs/}}, which uses non-leaky integrate-and-fire neurons with a linear response function. 
The sub-threshold neuron dynamics of the non-leaky integrate and fire neurons are described as follows:
\begin{eqnarray}
    \dot{v} &=& R \cdot \big(I_{syn}(t) + I_{bias}\big) \\
    I_{syn}(t) &=& W S(t)
\end{eqnarray}
where $v$ is the membrane potential of the neuron, $R$ is a constant, $I_{syn}$ is the synaptic input current, $I_{bias}$ is a constant input current term, $W$ is the synaptic weight matrix and $S(t)$ is a vector of input spike trains. For the results presented in this paper, we assume $R=1$ without any loss of generality. Upon reaching a spiking threshold $v_{th}$ the neuron's membrane potential is reduced by a value $v_{th}$ (not reset to zero). 

As a result of the above, between times $t$ and $t+\delta t$, for a total input current $I(t) = I_{syn}(t) + I_{bias}$, the neurons generate a number of spikes $n(t, t+\delta t)$ given by the following equation:
\begin{equation}
    n(t, t+\delta t) = \left \lfloor \frac{R \cdot I(t)\cdot \delta t}{v_{th}} \right \rfloor.
\end{equation}
In order to simulate the equivalent SNN model on Sinabs, the CNN's pre-trained weights are directly transferred to the equivalent SNN.


\subsection{Digit recognition on DVS recordings}

\begin{figure}
\begin{center}
\includegraphics[width=\textwidth]{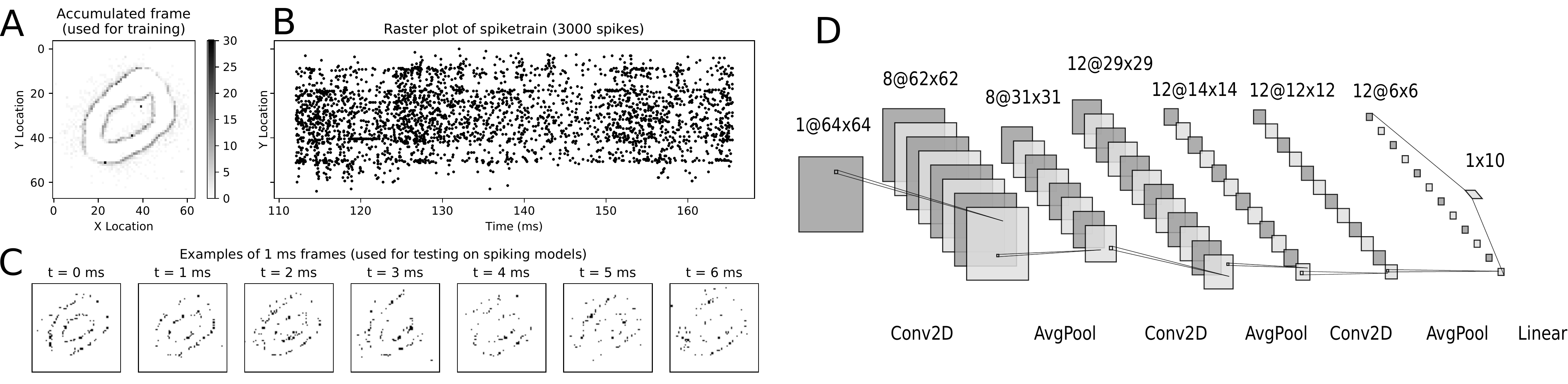}
\end{center}
\caption{Illustration of the MNIST-DVS dataset, as used in this work, and of the network model we used for the task. A: a single accumulated frame of 3000 spikes, as used for training. B: the corresponding 3000 spikes, in location and time. C: example single-millisecond frames, as sequentially shown to the Sinabs spiking network during the tests presented in figure \ref{fig:main}. D: The convolutional network model we used for this task. All convolutional layers and the linear layer are followed by ReLUs. Dropout is used before the linear layer at training time.}\label{fig:dataset}
\end{figure}

\subsubsection{Task and Dataset}
As a benchmark to assess the performance of the above training methods, we used an image recognition task on real data recorded by a Dynamic Vision Sensor (DVS). Given a spike train generated by the DVS, our spiking networks identify the class to which the object belongs---corresponding to the fastest-firing neuron in the output layer. For this task, we used the MNIST-DVS dataset at scale 16 \citep{serrano2015poker,liu2016benchmarking}, a collection of DVS recordings where digits from the classic MNIST dataset \citep{lecun1998mnist} are shown to the DVS camera as they move on a screen. 

During the training phase, we presented the (analog) neural network with images formed of accumulated DVS events, i.e. DVS spike trains divided into chunks and collapsed along the time dimension. The value of each pixel (0-255 in the image encoding we chose) was determined simply by the number of events on that pixel. The DVS recordings were split into frames not based on time length, but according to event count: the accumulation of each frame was stopped when the total number of events per frame reached a value of 3000 (figure \ref{fig:dataset}A). This ensured all frames had comparable pixel values without the need for normalization, and all contained similar amounts of information regardless of the type of activity presented. The information regarding event polarity was discarded, resulting in a 1-channel input frame (analogous to gray-scale image).

During testing on the spiking network simulation, the corresponding spike trains were presented to the network with 1 ms time resolution. This value was chosen to enable reasonable simulation times but could be lowered if needed. (Figure \ref{fig:dataset}C), to simulate the real-time event transmission between the DVS and a neuromorphic chip. 
The network state was reset between the presentation of a data chunk and the next.  
The polarity of events was ignored. Of the original 10000 recordings (1000 per digit from zero to nine), we set $20\%$ aside as test set.

\subsubsection{Network architecture}

In order to solve the task mentioned above, we used a simple convolutional neural network, with three 2D convolutional layers (3x3 filters), each followed by an average pooling layer (2x2 filters) and a rectified linear unit. The choice of average pooling is due to the difficulties of implementing max pooling in spiking networks \cite{rueckauer2017conversion}. The last layer is a linear (fully connected) layer, which outputs the class predictions (Figure \ref{fig:dataset}D). We used a cross-entropy loss function to evaluate the model predictions and optimized the network weights using the Adam optimizer \citep{kingma2014adam} with a learning rate of $10^{-3}$. Bias parameters were deactivated everywhere in the network. A 50\% dropout was used just before the fully connected layer at training time. The network was implemented using PyTorch \citep{paszke2017automatic}.

The whole procedure can be summarized as follows:
\begin{enumerate}
\item The dataset is prepared by dividing the original DVS recordings in sections of 3000 spikes each, ignoring event polarity. From these the following is saved:
    \begin{enumerate}
    \item The spike train itself, used for testing
    \item An image, corresponding to the time-collapsed spike train, with pixel values equal to the number of spikes at that location, used for training
    \end{enumerate}
\item A neural network is trained, applying quantization in correspondence with every ReLU. The loss used for training is binary cross-entropy with the addition of the synoploss term (eq. \ref{eq:synoploss}).
\item The trained weights are transferred to a spiking network simulation, implemented in Sinabs. The network dynamics is simulated with 1 ms time resolution. The network prediction is defined as the neuron that spikes the most over the 3000-spike input. Synaptic operations are counted as the sum of spikes emitted by each layer, weighted on the fan-out of that layer.
\end{enumerate}

For reproducibility, the python code implementing these methods is available at \url{gitlab.com/aiCTX/synoploss}.

\subsection{Object recognition on CIFAR-10}
\label{subsec:CIFAR10}

\subsubsection{Task and Dataset}
In order to validate the approach on a dataset with higher complexity than MNIST, we also benchmarked our work on CIFAR-10~\citep{krizhevsky2009learning}, a visual object classification task.
The input images were augmented with random crop and horizontal flip, and then normalized to $[-1, 1]$. A 20\% dropout rate was applied to the input layer to further augment the input data. 

For the experimental results on this dataset, we directly injected the image pixel analog value to the first layer of convolutions as input current in each simulation time step for $N_{dt}$ time steps. 
The magnitude of the current was scaled down by the same value $N_{dt}$, in order to have an accumulated current, over the whole simulation, equal to the analog input value. The Sinabs simulations were run for $N_{dt}=10$ time steps, obtaining SynOps and accuracy values.
The network state was reset between the presentation of an image and the next.


\subsubsection{Network architecture and training procedure}
In order to solve the task mentioned above, we used an All-ConvNet~\citep{springenberg2014striving}, a 9-layer convolutional network, without bias terms, which has 1.9M parameters in total. 
The ReLU layers in the model, including the last output layer, were replaced with QReLUs.
All the convolutional layers in this network are followed by a dropout layer with a rate of 10\%, which not only prevents over-fitting, but also compensates the SNN's discrete representations of analog values.
As illustrated in~\cite{springenberg2014striving}, training lasts 350 epochs, and the learning rate is initialized at $2.5 \times 10^{-4}$ and scaled down by a factor of 10 at epochs [200, 250, 300].
We use the Adam optimizer with weight decay of $10^{-3}$.
Note that the model was trained without ReLU on the last output layer, since it is harder to train the classification layer when the outputs are only positive, while the classification accuracy was tested with ReLU on the output layer, in order to have an equivalent network to the spiking model.

The entire experiment is as follows:
\begin{enumerate}

\item Train an ANN network, \textit{anet}, get its MAC and test the accuracy with original CIFAR10 dataset.
\item Scale up the weights of the first layer of \textit{anet} by $\rho$, and transfer the weights to the SNN equivalence, \textit{snet}.
    \begin{enumerate}
    \item test the accuracy and SynOps of \textit{snet} with $N_{dt}=10$, the input current is 1/10 of a pixel value.
    \item increase $\rho$, and repeat 2(a) till the accuracy reached about the ANN accuracy.
    \end{enumerate}
\item Select a $\rho$ from Step 2, where the \textit{snet} have SynOps $>$ MAC, and start quantization-aware training with the SynOp loss.
    \begin{enumerate}
    \item set the target SynOp to half of the current SynOps and train.
    \item test the accuracy and get the SynOps, then repeat 3(a) until the accuracy is too low to be meaningful, and thus a full accuracy/SynOps curve is obtained.
    \end{enumerate}
\end{enumerate}

\subsubsection{SynOps Optimization}
Before training the network with QReLU activations, the network was first trained with ReLU to get an initial set of parameters.
The network with QReLU was then initialized with the scaled parameters (scaling up by $\rho$ on the first layer).
The scaling factor $\rho$ was chosen to initialize the network in a state where enough information is propagated through layers so that the network performs reasonably well. Consequently, the weights of the last weighted layer were scaled by $1/\rho$, in order to adapt the classification loss back to its original range.

During testing, we measured the ANN and SNN performance in terms of their accuracy and SynOps, and found a mismatch of SynOps between training and testing.
There are two main reasons:
1) The output of a dropout layer (with a dropout rate $p$) is always scaled down by $1-p$ to compensate the dropped out activations, however the mismatch could be large after a sequence of dropout layers.
2) Due to discrete spike events operated in the network where the order (not only the count) of the spikes matters, the mismatch occurs between the spike count-based analog activation and the actual spiking ones.

To compensate for this mismatch, for all the trained models we tested the performance with both $1.5\times$ and $2\times$ scaled-up first layer weights.
Lastly, we optimized the QReLU-based model with the objective of minimizing the classification error given a target SynOps.
We trained 30 models with lower and lower target SynOps, and each model was initialized with the trained weights of the previous one.

\section{Results}

\subsection{The SynOp loss term leads to a reduction in network activity on DVS data}

\begin{figure}
\begin{center}
\includegraphics[width=\textwidth]{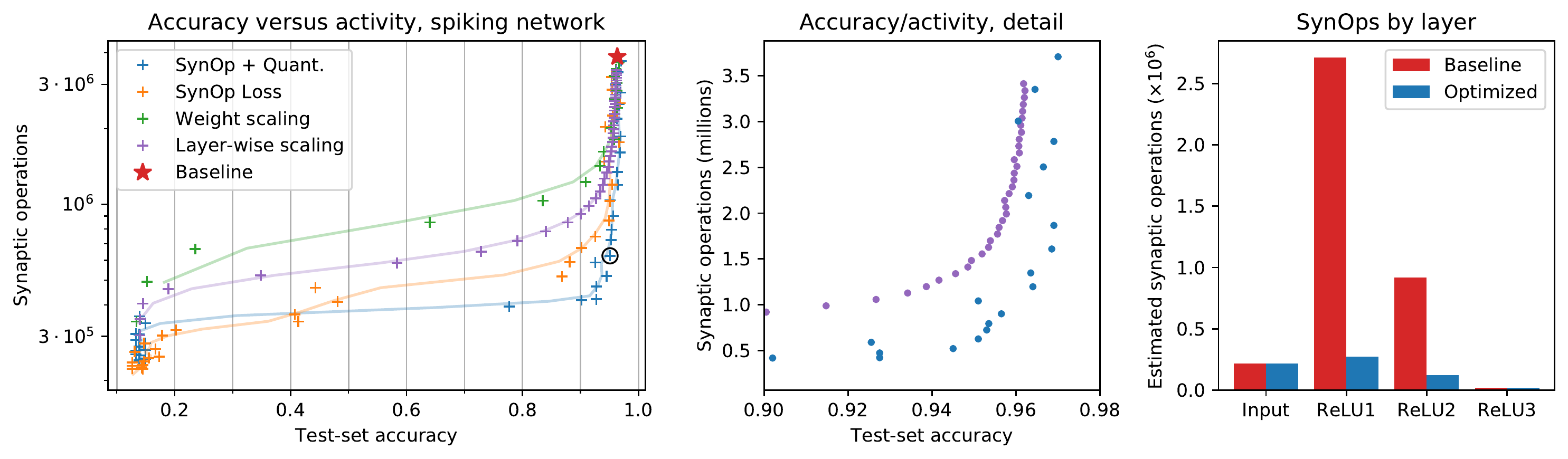}
\end{center}
\caption{Results on the MNIST-DVS dataset. Left: SynOps-accuracy curves computed on a spiking network simulation with Sinabs. Each point represents a different model, trained for a different value of $S_0$ or rescaled by a different value of $\rho$.  The red star represents the original model, standard CNN weights transferred to the SNN without changes. The solid lines are smoothed versions of the curves described by the data points, provided as a guide to the eye. Center: a zoomed-in version of the left panel, showing direct comparison between layer-wise gain scaling \citep{rueckauer2017conversion} and our method. Right: SynOps per layer, compared between the baseline model and a selected model trained with the SynOp loss and quantization. This is the model indicated by a black circle in the first panel. Note that the input SynOps depend only on the input data, and cannot be changed by training.}\label{fig:main}
\end{figure}

In figure \ref{fig:main}, we show the results of four methods to reduce the activity of the network, in a way that yields energy savings. First, as a baseline, we trained a traditional CNN using a cross-entropy loss function, and rescaled down the weights of its first layer. This is equivalent to rescaling the input values, and has the effect of proportionally reducing the activity in all subsequent layers of the network. The ``baseline'' model in figure \ref{fig:main} is the same network, with no input rescaling: weights are transferred from the CNN to the corresponding layers of the SNN without any changes or special considerations. Thresholds are set to 1 on all layers.
Second, following \cite{rueckauer2017conversion}, we rescaled the weights of each layer in such a way that the maximum activation in each layer stays constant (see Methods). The input weights are again rescaled as stated above.
Third, we introduced an additional term in the loss function, the SynOp loss, which directly pushes the estimated number of SynOps to a given value. We trained CNN models, each with a different target number of synaptic operations, independently of each other. 
Furthermore, excessive reduction of the SynOps leads to the silencing of certain neurons, and other discretization errors, causing an immediate drop in accuracy. To account for this we jointly use the SynOp loss term and quantization-aware training.

We tested our training methods on a real-world use case of spiking neural networks. Dynamic Vision Sensors (DVS) are used in neuromorphic engineering as very-low-power sources of visual information, and are a natural data source for SNNs simulated on neuromorphic hardware. We transferred the weights learned with the methods described above onto a spiking network simulation, and used it to identify the digits presented to the DVS in the MNIST-DVS dataset.

Our results show that adding a requirement on the number of synaptic operations to the loss yields better results in terms of accuracy compared to rescaling input weights and layer-wise activation gains (Figure \ref{fig:main}, orange). Using the SynOp loss together with quantization during training outperforms the simpler methods, allowing for further reduction of the SynOps value with smaller losses in accuracy (Figure \ref{fig:main}, blue).

Among the models trained in this way, we selected one with a good balance between energy consumption and accuracy, and used it for a direct comparison with the baseline (that is, weights from an ANN without quantization and no additional loss terms). The second and third panels of figure \ref{fig:main} graphically show the large decrease in the number of synaptic operations required by each layer of our model, and the very small reduction in performance. This particular model brings accuracy down from 96.3\% to 95.0\%, while reducing the number of synaptic operations from 3.86M to 0.63M, an 84\% reduction of the SynOp-related energy consumption.

\subsection{The SynOp loss leads to a lower operations count compared to ANNs on CIFAR10}

\begin{figure}
\begin{center}
\includegraphics[width=1.\textwidth]{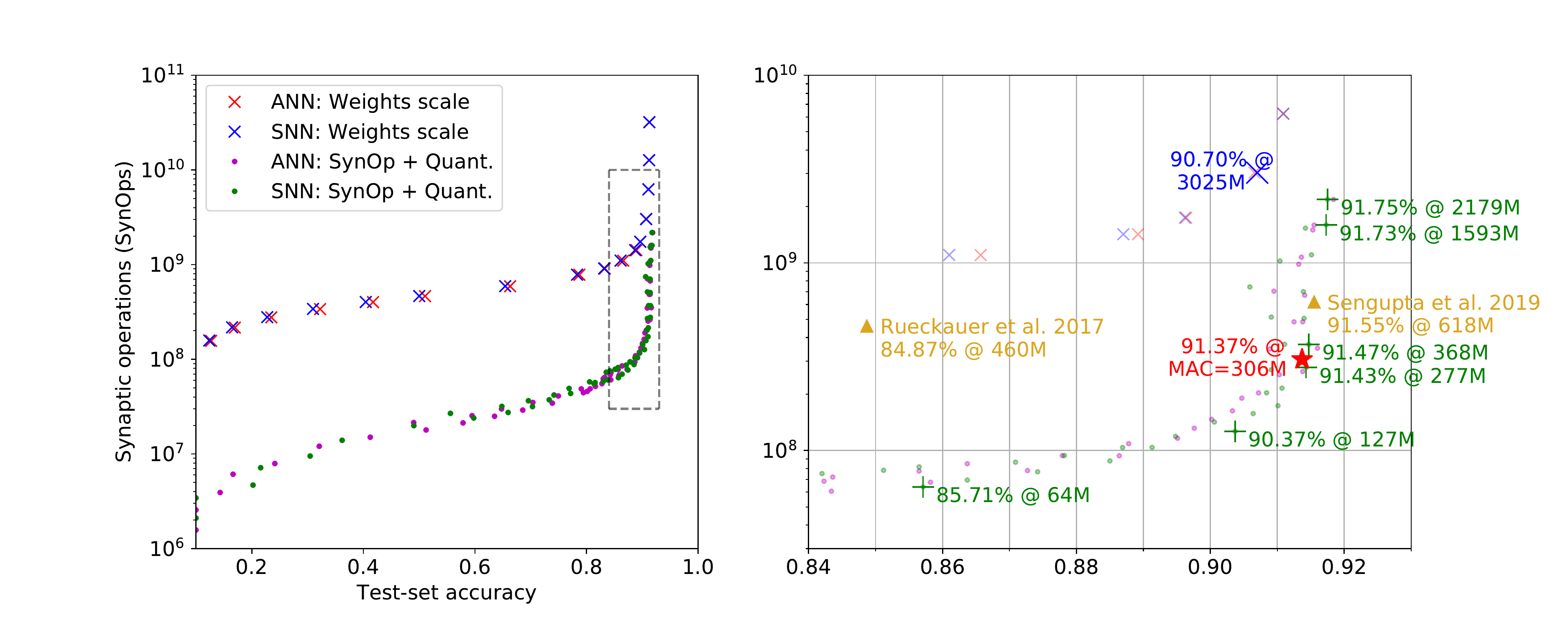}
\end{center}
\caption{Accuracy vs.\ SynOps curves on the non-spiking CIFAR-10 task.
    `ANN' results (red crosses and purple dots) are SynOp count estimations based on the quantized activations of an analogue network. `SNN' results (blue crosses and green dots) are SynOp values obtained by Sinabs simulation.
    The performance of models fine-tuned with SynOp loss and quantization (dots) shows a clear advantage over weight rescaling (crosses).
    Right panel: a zoomed-in plot of the dashed region.
    A high-light blue cross represents a good performance of the arbitrary weight scaling.
    The good trade-off points, trained with SynOp loss and quantization, listed in table \ref{tbl:snn_compare} are marked with green `+'.
    The results from other work are marked with orange triangles. 
    The original ANN model, for which the MAC is plotted instead of the SynOps, is marked with a red star.
}\label{fig:act-acc-cifar}
\end{figure}

SNNs are a natural way of working with DVS events, having advantages over ANNs in event-driven processing. 
However, it is also interesting to highlight the benefits of using SNNs over ANNs in conventional non-spiking computer vision tasks, e.g. CIFAR-10, where SNNs can still offer advantages in power consumption.
As stated in Section~\ref{subsec:CIFAR10} we have trained the network with two approaches:
1) conventional ANN training plus weight scaling as the baseline;
2) further training with QReLU and SynOp Loss for performance optimization.

\subsubsection{Weight Scaling}
We first trained the analog All-ConvNet on CIFAR-10, attaining an accuracy of 91.37\% and a MAC of 306M (red star in Figure~\ref{fig:act-acc-cifar}).
Then, we transferred the trained weights directly on the equivalent SNN and scaled the weights of its first layer to manipulate the overall activity level.
This is shown by the blue crosses in Figure \ref{fig:act-acc-cifar}: as the SynOp count grows, so does the accuracy.
However, the SynOps are around 10 times to the MAC of ANN when the accuracy reaches an acceptable rate of 90.7\%.
To improve on this result, we fine-tuned this training by adding quantization and the SynOp loss. 

A faster way to measure the same quantities is by testing the analog model, with ReLU layers all replaced with QReLU, and count the activation levels instead of the Sinabs spike counts.
Estimations based on this quantized activation layers are shown as red crosses in Figure~\ref{fig:act-acc-cifar}.
The performance on accuracy and SynOps of the analog network and its spiking equivalent are well aligned, showing that quantized activations are a good proxy for the firing rates of the simulated SNN, at least in this regime.

\subsubsection{SynOp loss optimization}

We further fine-tuned one of the weight-scaled models obtained above, with the addition of quantization-aware training and the SynOp loss.
Figure~\ref{fig:act-acc-cifar} also shows the classification accuracy and SynOps for both quantized-analog and spiking models (blue and green dots respectively) trained with this method.

Multiple SNN test trials achieve better accuracy than the original ANN model~(red star, 91.37\%), thanks to the further training with QReLU.
As the SynOp goes down, the accuracy stays above the original ANN model until 91.43\% when SynOps are at 277M~(see one of the green `+' in Figure~\ref{fig:act-acc-cifar}).
Note that, the SNN has outperformed ANN both on accuracy and operations count, where the number of MAC in the original ANN is 306M.
As another good example of accuracy-SynOp trade-off (90.37\% at 127M), our model could perform reasonably well, above 90\%, by reducing 58\% (Syn-MAC ratio is 0.42) of computing operations from the original ANN.
Therefore, running the SNN model on neuromorphic hardware will benefit on energy efficiency not only from the lower computation cost of SynOps but also from the significant reduction on operation counts. 
Additionally, the plot shows how this method outperforms weight scaling in terms of operation counts by roughly a factor of 10 for all accuracy values.

As far as we know, our converted SNN model from the AllConvNet reached the state-of-the-art accuracy at 91.75\% among SNN models (see detailed comparison in Table~\ref{tbl:snn_compare} and Figure~\ref{fig:act-acc-cifar}).
In addition, our model is the smallest, at 1.9M parameters, while the BinaryConnect model~\citep{rueckauer2017conversion} is 7 times larger in size and WeightNorm, consisting of a VGG-16~\citep{sengupta2019going}, is 8-fold in size.
Although achieving the best accuracy requires a SynOp of 2,179M, this can easily be reduced by 27\% by giving up 0.02\% in accuracy, see the two green `+' on the top-right of Figure~\ref{fig:act-acc-cifar}.
Comparing to the result from \cite{sengupta2019going} (orange triangle on the right of Figure~\ref{fig:act-acc-cifar}), our model achieves 91.47\% in accuracy at 368M SynOps, thus only loses 0.08\% in accuracy but saves 41\% of SynOps and energy.
Thanks to the optimization of the SynOp loss, the number of SynOps is continuously pushed down while keeping an appropriate accuracy, e.g. 85.71\% at a SynOp of 64M.
This result not only outperforms most of the early attempts of SNN models for the CIFAR-10 task~\citep{panda2016unsupervised,cao2015spiking,hunsberger2015spiking}, but also brings down the SynOps to only 1/5 of the MAC and saves 86\% energy compared to \cite{rueckauer2017conversion}. 

In a brief summary, 1) the energy-aware training strategy pushes down the SynOps 10 times compared to its weight scaling baseline; 2) the QReLU-trained SNN achieves the state-of-the-art accuracy in CIFAR-10 task; and 3) the trade-off performances between accuracy and energy show a significant save in computation cost/energy comparing to existing SNN models and the equivalent non-spiking CNN.

\begin{table}
\centering
\begin{tabular}{|c|cc|ccc|ccc|}
\hline
SNN Models    & \multicolumn{2}{c|}{Net Architecture}       & \multicolumn{3}{c|}{Best Accuracy }            & \multicolumn{3}{c|}{Accuracy-SynOps trade-off}  \\ 

\cline{2-9}

       & N. par.     & MAC           & Acc.       & SynOps        &\begin{tabular}[c]{@{}c@{}}Syn-MAC\\ratio\\\end{tabular}     & Acc.      & SynOps      &\begin{tabular}[c]{@{}c@{}}Syn-MAC\\ratio\\\end{tabular}     \\ 
\hline

BinaryConnect~  & 14M           & 616M          & 90.85          & N/A           & N/A           & \textcolor{blue}{84.87}        & \textcolor{blue}{460M}   & \textcolor{blue}{0.75}      \\
\hline
WeightNorm~         & 15M           & 313M          & 91.55          & \textbf{618M} & \textbf{1.98} & \textcolor{ForestGreen}{\textbf{91.55}}   & \textcolor{ForestGreen}{618M}     & \textcolor{ForestGreen}{1.98}   \\

\hline
Ours            & \textbf{1.9M} & \textbf{306M} & \begin{tabular}[c]{@{}l@{}}\textbf{91.75}\\91.73\\\end{tabular} & \begin{tabular}[c]{@{}l@{}}2179M\\1593M\\\end{tabular} & \begin{tabular}[c]{@{}l@{}}7.12\\5.21\\\end{tabular} & \begin{tabular}[c]{@{}l@{}}\textcolor{ForestGreen}{91.47}\\91.43\\90.37\\\textcolor{blue}{\textbf{85.71}}\end{tabular} & \begin{tabular}[c]{@{}l@{}}\textcolor{ForestGreen}{\textbf{368M}}\\277M\\127M\\\textcolor{blue}{\textbf{64M}}\end{tabular} & \begin{tabular}[c]{@{}l@{}}\textcolor{ForestGreen}{\textbf{1.20}}\\0.91\\0.42\\\textcolor{blue}{\textbf{0.21}}\end{tabular}  \\
\hline
\end{tabular}\caption{Comparison with best SNN models on CIFAR-10. BinaryConnect: 8-layer ConvNet from \cite{rueckauer2017conversion}; WeightNorm: VGG-16 model from \cite{sengupta2019going}.
The most efficient model and the best performance are highlighted as bold fonts.
Regarding to the comparisons on accuracy-SynOps trade-off, blue colored result in our models refers to the performance close to that of \cite{rueckauer2017conversion}; while green shows the result approximates to \cite{sengupta2019going}.
And the numbers in bold only highlight the winner in these two comparisons.
}
\label{tbl:snn_compare}
\end{table}

\subsubsection{SynOp vs. Accuracy for shorter inference times}
\begin{figure}
\begin{center}
\includegraphics[width=.49\textwidth]{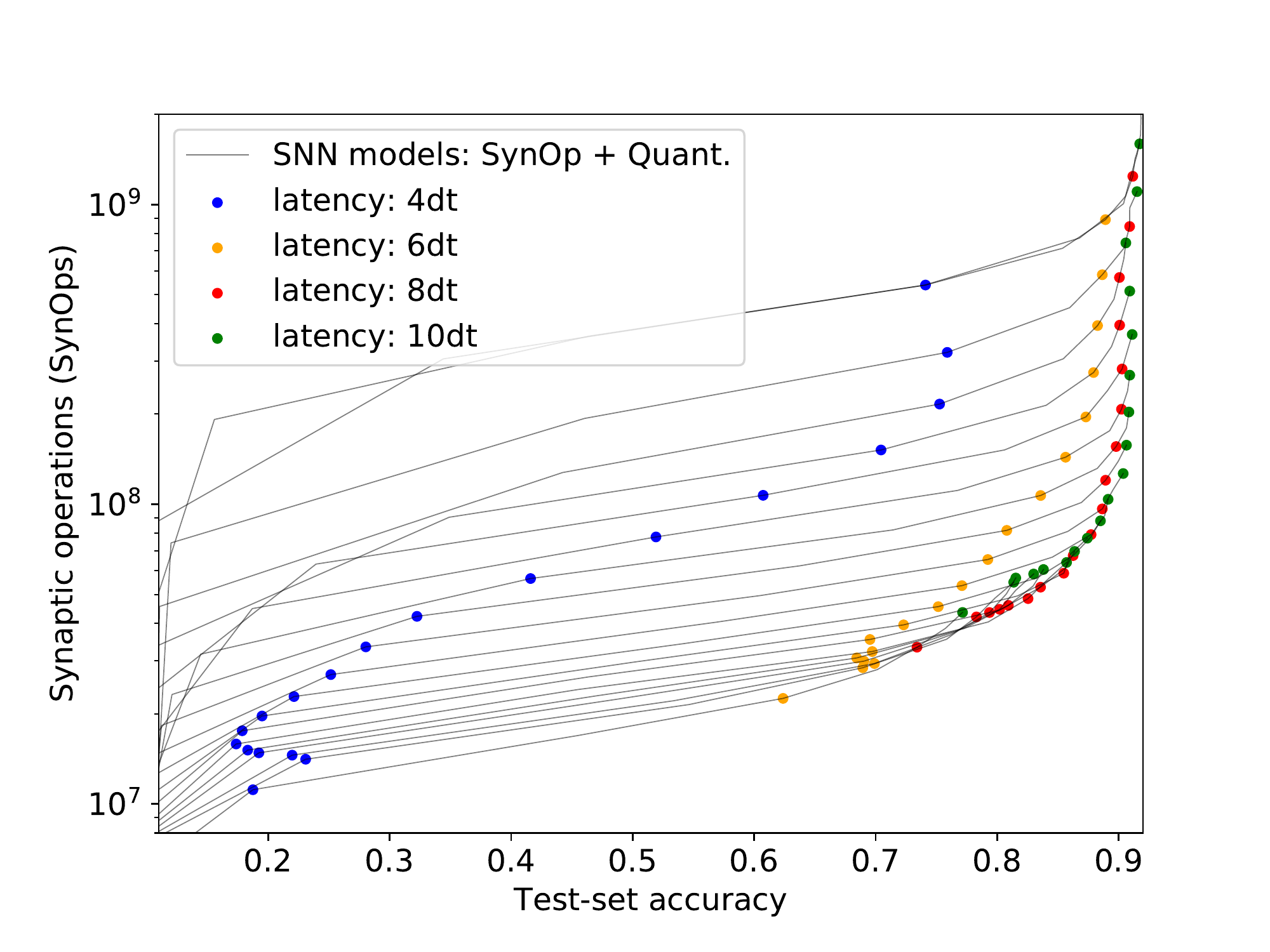}
\end{center}
\caption{Total activity and accuracy on the CIFAR-10 benchmark for increasing inference times (current injected for longer times, thus leading to more SynOps). Gray lines correspond to individual SNN models, tested at $N_{dt} = 1,...,10$. To facilitate same $N_{dt}$ comparison, coloured dots were added for all models at $N_{dt} = 4, 6, 8, 10$.}
\label{fig:latency-cifar}
\end{figure}

Unlike the DVS data, which has its own time dynamics, there are no restrictions on how static images should be presented to the network in time.
Therefore we measure total spike count instead of firing rates, thus calculating the total energy cost per image, independently of time.
For example, Figure~\ref{fig:act-acc-cifar} shows how SynOp loss optimization pushes the SynOp to a value lower than the MAC during training.
This is one of the approaches in which SNNs outperform ANNs in the accuracy-operations trade-off; while the other benefit SNNs naturally bring is the temporal encoding and computation.
SNNs continuously output a prediction from the moment when the input currents are injected. This prediction becomes more accurate with more time.
For the experiments presented in the previous sections, we only measure the classification accuracy when the input is completely forwarded into the network, $N_{dt} = 10$: the input currents are chosen so that the total input accumulated over $N_{dt} = 10$ time steps is equivalent to that of the analog network during training.
In Figure~\ref{fig:latency-cifar}, we measure how SNN models perform in the course of the entire process, $N_{dt} = 1, 2, 3, ..., 10$. The figure shows how classification accuracy increases when more simulation time-steps are allowed, and therefore more accumulated current is injected to the network.
Each gray curve represents a single trained model, and the SynOp and accuracy are tested with increasing $N_{dt}$.
The colored dots mark the accuracy-SynOps pair at $N_{dt}=4,6,8,10$ over all trained networks.
The same-colored dots approach to the expected SNN result (green dots at $N_{dt}=10$) as $N_{dt}$ increases.

\begin{figure}
\begin{center}
\includegraphics[width=.65\textwidth]{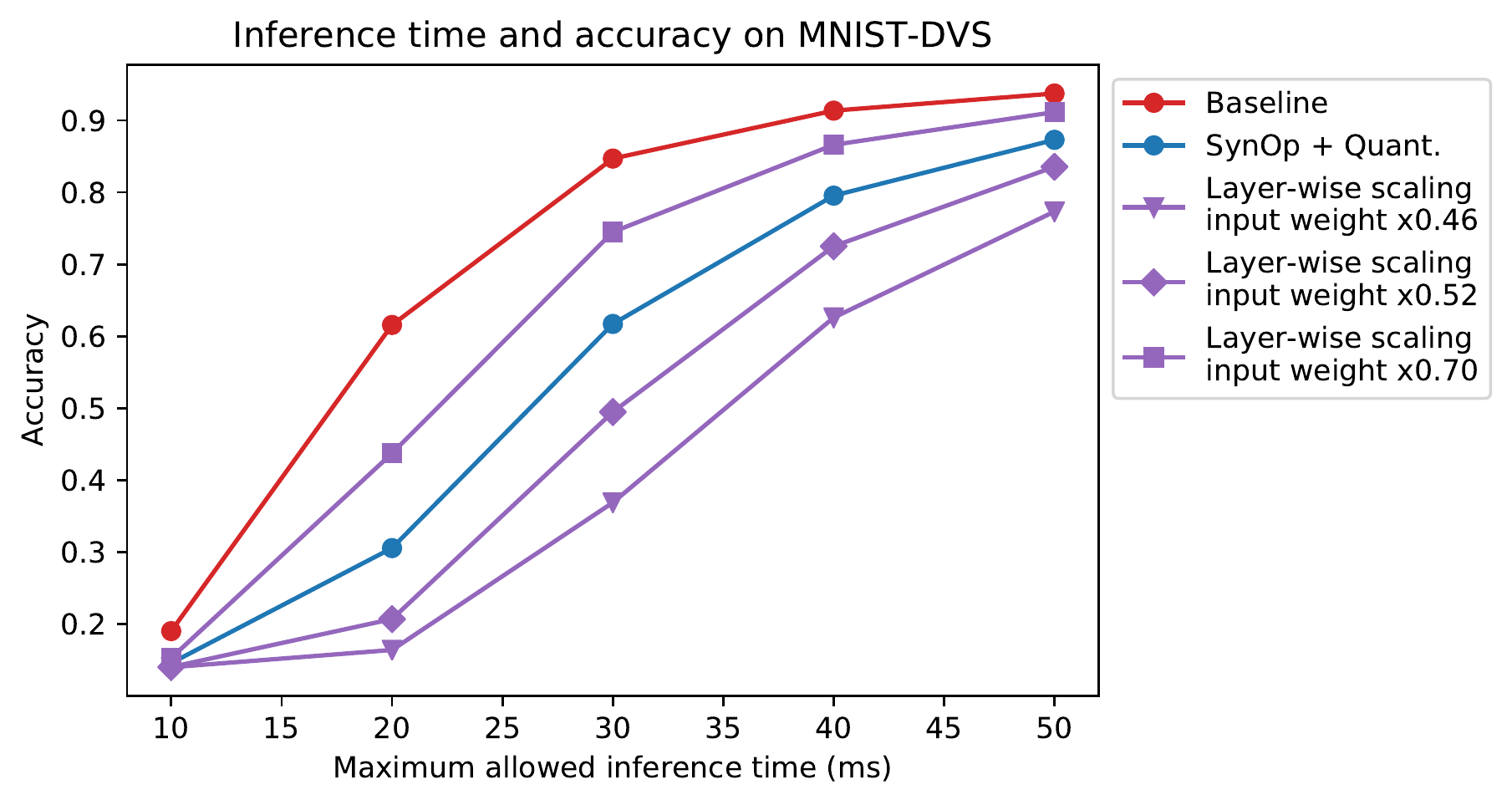}
\end{center}
\caption{Limited-time inference on MNIST-DVS. Here, the accuracy of the networks is measured at a limited input length of 10, 20, 30, 40, 50 ms. The accuracy of the network trained with our method (again, we chose the one indicated by the black circle in figure \ref{fig:main}) behaves, in relation to observation time, similarly to that of other networks, but with lower power consumption. The models and color choices are the same as in figure \ref{fig:main}. The different purple lines correspond to different rescaling of the input layer weights, chosen so to have a comparable accuracy with the other curves.}
\label{fig:latency-dvs}
\end{figure}

On the other hand, understanding the relationship between inference time and accuracy is very relevant when dealing with DVS data. In general, a global reduction of spike rates in a rate-based network causes a corresponding increase in the latency, since more time is needed to accumulate enough spikes for a reliable prediction. We compared one of our networks with an equivalent model prepared through the `robust' layer-wise normalization technique from \cite{rueckauer2017conversion}, with a few different values of input weight scale. Figure \ref{fig:latency-dvs} shows that the dependency of accuracy on the inference time follows a similar trajectory for all these models. We conclude that the increase in latency does not depend on the specific method used for optimization, and our network's latency is similar to that of other models with similar accuracy, despite the much lower power consumption (shown in previous sections).

\subsubsection{Effects on weight statistics}
\begin{figure}
\begin{center}
\includegraphics[width=.9\textwidth]{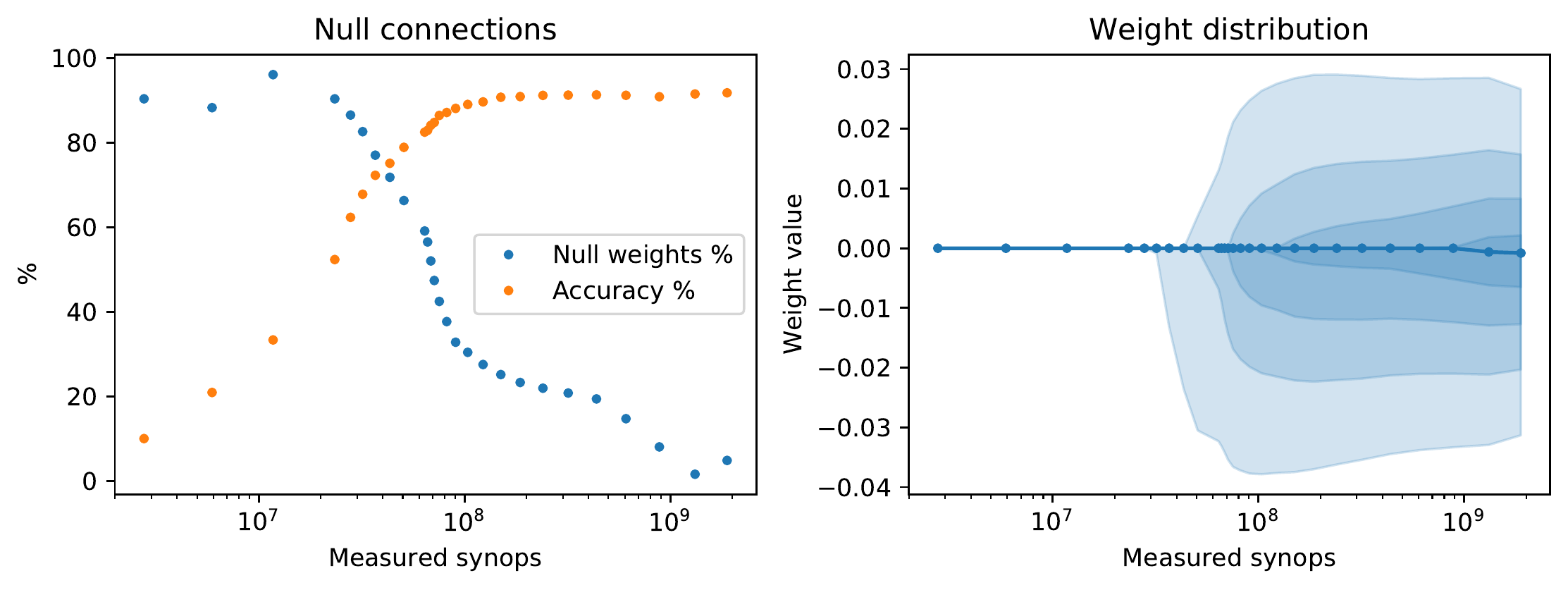}
\end{center}
\caption{Effect of the SynOp loss on the network's weights. Left: the fraction of near-zero weights ($|w| < 10^{-9}$) greatly increases in models where a stricter reduction of SynOp counts were imposed. The test-set accuracy values for each model are also shown for comparison. Right: the distribution of weights as a function of the SynOp count. Shaded areas indicate, from lighter to darker, the following inter-quantile ranges: 10\%--90\%; 20\%--80\%; 30\%--70\%; 40\%--60\%. The solid line is the median weight. The  models used for this test are the same as those shown in figure \ref{fig:act-acc-cifar}, trained with quantization on static CIFAR images.}
\label{fig:pruning-cifar}
\end{figure}

Common regularization techniques in ordinary neural networks often involve the inclusion of an $L_1$ or $L_2$ cost on the network weights. In rough, intuitive terms, $L_1$ regularization has a sparsifying effect, pushing smaller connections towards zero; $L_2$ regularization generally keeps the weights from growing to excessively large values. Conversely, the effect on weights of penalizing synaptic operations or reducing the network's activity, as we do with the SynOp loss term, is not immediately clear. We investigate whether imposing low synaptic operations count has a sparsifying effect on the weight structure. To this end, we examine how many synaptic connections in our models are null connections, which we define as weights $w$ such that $|w| < 10^{-9}$ (this threshold can be changed by several orders of magnitude without impacting the conclusions). We performed this analysis on the networks trained on the CIFAR-10 dataset, as explained in the previous sections. These networks are much wider and deeper than the ones used for the MNIST-DVS task, and therefore can better show weight sparseness effects. Figure \ref{fig:pruning-cifar} (left) shows how the fraction of null weights changes with the SynOp count (and thus, of the regularization strength), and compares it with the model's accuracy. When the number of synaptic operations is forced to be extremely low, the fraction of null weights reaches values above 90\%. A large increase in null connections, however, is already noticeable for models above 80\% accuracy, showing that the SynOp loss term does have a sparsifying effect, and that this is desirable. For the sake of completeness, in figure \ref{fig:pruning-cifar} (right), we also show a depiction of the distribution of weights as a function of the number of synaptic operations.

Setting synaptic weights to zero is effectively equivalent to pruning certain connections between a layer and the next. Other than $L_1$ regularization of the weights, more sophisticated pruning-and-retraining algorithms have been studied in the machine learning literature \citep{lecun1990optimal, hassibi1993second}. However, advanced pruning methods (such as those based on Fisher information) are usually coupled with partial retraining of the network, and are therefore more alike a form of architectural search \citep{crowley2018pruning}. Due to the retraining of the remaining weights, these forms of pruning are not guaranteed to reduce the activity levels if not coupled to other forms of regularization.

\section{Discussion and Conclusion}

We used two techniques which significantly improve the energy requirements of machine learning models that run on neuromorphic hardware, while maintaining similar performances.

The first improvement consisted in optimising the energy expenditure by directly adding it to the loss function during training. This method encourages smaller activations in all neurons, which is not in itself an issue in analog models, but can lead to discretization errors, due to the lower firing rates, once the weights are transferred to a spiking network. To solve this problem, we introduced the second improvement; quantization-aware training, whereby the network activity is quantized at each layer, i.e. only integer activations are allowed. Discretizing the network’s activity would normally reduce all gradients to zero: this can be solved by substituting the true gradient with a surrogate.

Applying these two methods together, we achieved an up to ten-fold drop in the number of synaptic operations and the consequent energy consumption in the DVS-MNIST task, with only a minor (1-2\%) loss in performance, when comparing to simply transferring the weights from a trained CNN to a spiking network. To demonstrate the scalability of this approach, we also show that, as the network grows bigger to solve a much more complex task of CIFAR-10 image classification, the SynOps are reduced to 42\% of the MAC, while losing 1\% of accuracy (90.37\% at 127M). The accuracy-energy trade-off can be flexibly tuned at training time. We also showed the consequences of using this method on the distribution of network weights and the network's accuracy as a function of time.



While training based on static frames is not the optimal approach to leverage all the benefits of spike-based computation, it enables fast training with the use of state-of-the-art deep learning tools. In addition, the hybrid strategy to train SNNs based on a target power metric is unique to SNNs. Conversely, optimizing the energy requirement of an ANN/CNN requires modification of the network architecture itself, which can require large amounts of computational resources \citep{cai2018proxylessnas}. In this work, we demonstrated that we can train an SNN to a target energy level without a need to alter the network hyperparameters. A potential drawback of this approach of (re)training the model as opposed to simply transferring the weights of a pre-trained model is brought to light when attempting to convert very deep networks trained over large datasets such as IMAGENET. Pre-trained deep CNNs trained over large datasets are readily available on the web and can be used to quickly instantiate a spiking CNN. The task becomes much more cumbersome to optimize for power utilization using the method described in this paper, ie. one has to retrain the network over the relevant dataset for optimal performance. However, our method can also be effectively used to fine-tune a pre-trained network, removing the need for training from scratch (Supplementary Figure 1). Furthermore, no large event-based datasets of the magnitude of IMAGENET exist currently, and perhaps when such datasets are generated, the corresponding models optimized for spiking CNNs will also be developed and made readily available.

The quantization and SynOp-based optimization used in this paper can potentially be applied, beyond the method illustrated here, in more general contexts such as algorithms based on back-propagation through time to reduce power usage.
Such a reduction in power usage can make a large difference when the model is ran on a mobile, battery-powered, neuromorphic device, with potential for a significant impact in the industrial applications.

\section*{Conflict of Interest Statement}
All authors were employed by aiCTX AG during the course of the work published in this article. 


\section*{Author Contributions}

SS designed research; QL and SS contributed to the methods; MS, QL, and MB contributed code and performed experiments; all authors wrote the paper.

\section*{Funding}

This work is supported in part by H2020 ECSEL grant TEMPO (826655). The funder was not involved in the study design, collection, analysis, interpretation of data, the writing of this article or the decision to submit it for publication.

\section*{Acknowledgments}
The authors would like to thank Mr. Felix Bauer, Mr. Ole Richter, Dr. Dylan Muir and Dr. Ning Qiao for their support and feedback on this work.


\section*{Data and Code Availability}
The third-party datasets used in this study are available from their respective authors, cited in the main text. The Python/PyTorch code used for training and analysis is publicly available at \url{gitlab.com/aiCTX/synoploss}. Reuse and feedback are encouraged, within the terms of the license provided.

\bibliographystyle{frontiersinSCNS_ENG_HUMS.bst} 
\bibliography{test}






\end{document}


\section*{\textit{Optimizing the energy consumption of spiking neural networks for neuromorphic applications}: Supplementary material}
 Martino Sorbaro, Qian Liu, Massimo Bortone and Sadique Sheik
 
 \vspace{2em}
 
 \subsection*{Supplementary figure 1}
 
 This figure includes the results of additional experiments, which use a larger network (VGG16: Simonyan and Zisserman, 2014), starting from a pretrained model, fine-tuned using the SynOp loss. The task is a more difficult image recognition problem, on a 10-class subset of ImageNet called \emph{ImageNette} (Howard, 2019: https://github.com/fastai/imagenette), which uses much larger images compared to CIFAR10. We also ran these experiments using a simple $L_1$ penalty on activations, without per-layer weighting based on layer fanout. We draw three conclusions:
 
 \begin{enumerate}
     \item SynOp loss training can be effectively used as a final training step on pre-trained networks, removing the need for training from scratch.
     \item The methods we illustrate are still effective when dealing with more difficult tasks and larger networks.
     \item There is no evidence that fanout-weighting is necessary, as an unweighted $L_1$ penalty seems to lead to similar results, at least in this limited test. Weighting the penalty according to layer fanout is still our preferred choice, since it is a more direct proxy of power consumption -- and the results may be different in other cases not evaluated here.
 \end{enumerate}
 
 \begin{figure}[h!]
     \centering
     \includegraphics[scale=0.7]{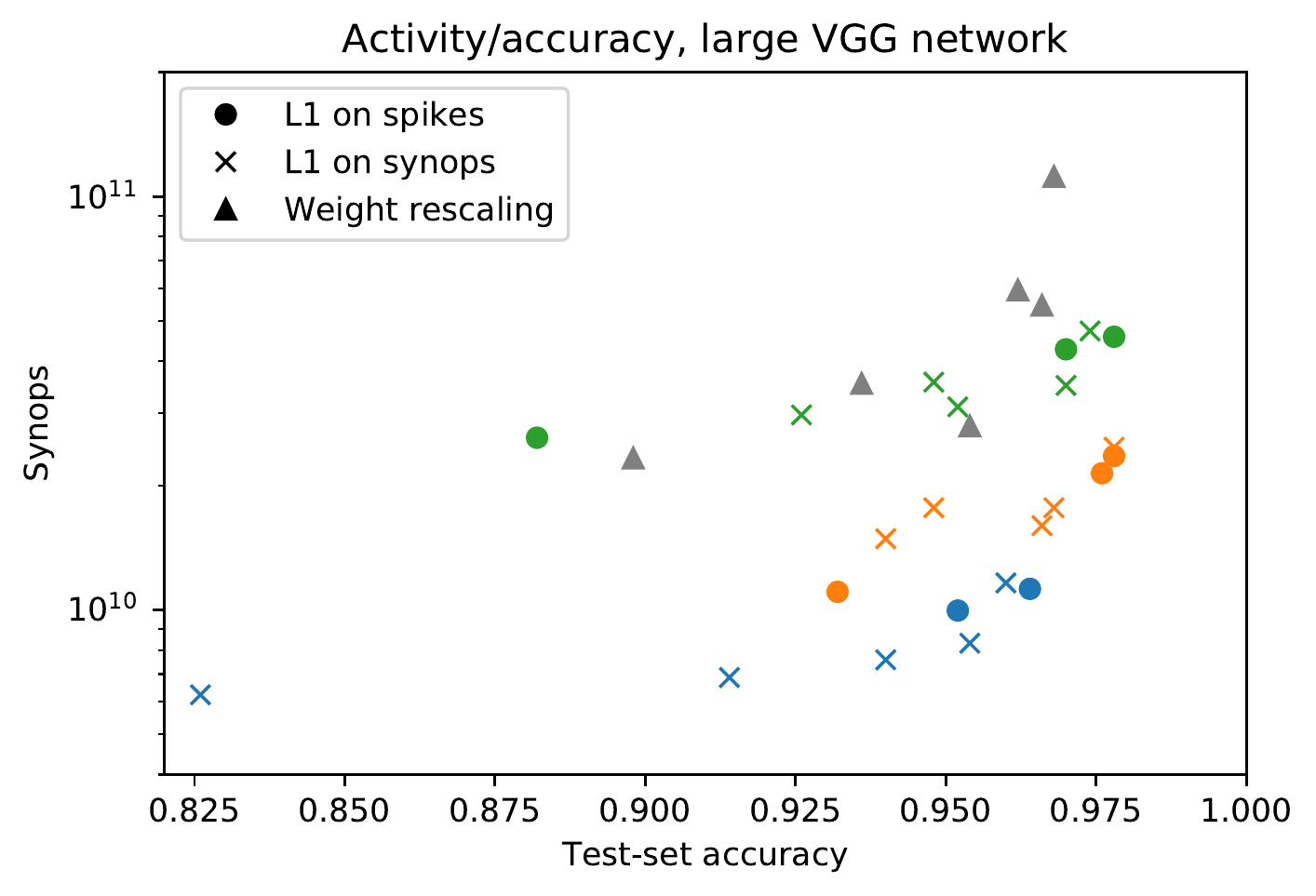}
     \caption{Activity/accuracy results for a VGG16 pre-trained network, fine-tuned using the SynOp loss. Colors correspond to different choices of rescaling of the input layer weights (causing increase or reduction of the overall activity). Crosses represent models trained with an $L_1$ penalty on the SynOp value (i.e. with fanout-weighting of the activations in the penalty). Circles represent models trained with a penalty on the number of spikes (non-weighted activations). Triangles are models trained with no additional loss term, for various values of input weight scaling.}
 \end{figure}

 \subsection*{Supplementary figure 2}
 
 In the SynOp loss penalty, we use an estimate of synaptic operations count, which is based on the quantized activations of the analog neural network. In this figure, we show that the value estimated in this way closely corresponds to the SynOp count actually observed in the simulated SNN with the same weights. Additionally there is also a good correspondence between the accuracy of the ANN and that of its associated SNN, limited to the regimes and networks used in this work. The models shown here are the same as those shown in figure 2 of the main text, with the same training methods, dataset and parameters.
 
  \begin{figure}[h!]
     \centering
     \includegraphics[scale=0.7]{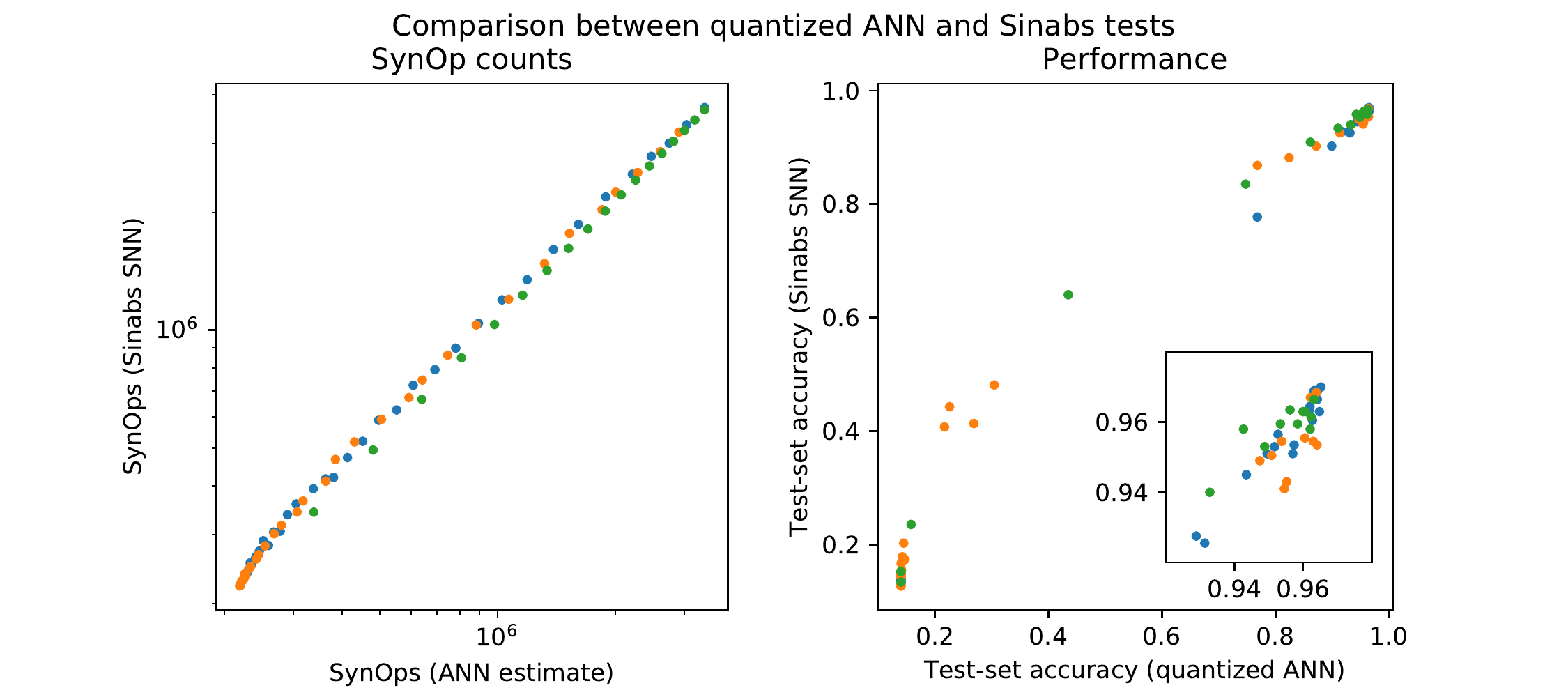}
     \caption{Left: comparison between the SynOps values as measured in our simulated spiking networks (from spike counts) and the SynOps estimated from the ANN (from quantized activations). Right: comparison between the accuracy of a quantized ANN versus the accuracy of the simulated SNN with the same weights. The inset shows the top-right area in more detail. The colors correspond to different training methods, with the same color scheme used in figure 2 of the main text.}
 \end{figure}